\documentclass[conference]{IEEEtran}
\IEEEoverridecommandlockouts
\usepackage{cite}
\usepackage{amsmath,amssymb,amsfonts}
\usepackage{algorithmic}
\usepackage{graphicx}
\usepackage{textcomp}
\usepackage{xcolor}
\usepackage{multirow}
\usepackage{bbm}

\def\BibTeX{{\rm B\kern-.05em{\sc i\kern-.025em b}\kern-.08em
    T\kern-.1667em\lower.7ex\hbox{E}\kern-.125emX}}
\begin{document}

\title{randomHAR: Improving Ensemble Deep Learners for Human Activity Recognition with Sensor Selection and Reinforcement Learning}

\author{\IEEEauthorblockN{1\textsuperscript{st} Yiran Huang, 2\textsuperscript{nd} Yexu Zhou, 3\textsuperscript{rd} Till Riedel, 4\textsuperscript{th} Likun Fang, 5\textsuperscript{th} Michael Beigl}
\IEEEauthorblockA{\textit{Pervasive Computing Systems-TECO}\\
\IEEEauthorblockA{\textit{Karlsruhe Institute of Technology}}
Karlsruhe, Germany \\
\{yhuang, zhou, riedel, fang, beigl\}@teco.edu}}


\maketitle

\begin{abstract}
Deep learning has proven to be an effective approach in the field of Human activity recognition (HAR), outperforming other architectures that require manual feature engineering. Despite recent advancements, challenges inherent to HAR data, such as noisy data, intra-class variability and inter-class similarity, remain. To address these challenges, we propose an ensemble method, called randomHAR. The general idea behind randomHAR is training a series of deep learning models with the same architecture on randomly selected sensor data from the given dataset. Besides, an agent is trained with the reinforcement learning algorithm to identify the optimal subset of the trained models that are utilized for runtime prediction. In contrast to existing work, this approach optimizes the ensemble process rather than the architecture of the constituent models. To assess the performance of the approach, we compare it against two HAR algorithms, including the current state of the art, on six HAR benchmark datasets. The result of the experiment demonstrates that the proposed approach outperforms the state-of-the-art method, ensembleLSTM.
\end{abstract}


\section{Introduction}
Human Activity Recognition (HAR) is a field that infers human activities from raw time-series signals acquired through embedded sensors of smartphones and wearable devices~\cite{ramanujam2021human}. It has emerged as a revolutionary technology for real-time and autonomous monitoring in behavior analysis, ambient assisted living, activity of daily living (ADL), elderly care, rehabilitations, entertainments, and surveillance in smart home environments~\cite{nweke2019data}.
Deep learning-based approaches, because of their automatic feature extraction capabilities, have been widely adopted~\cite{wang2019deep}. Recent studies have shown that Deep Learning approaches outperform classical Machine Learning algorithms in many HAR tasks~\cite{bock2021improving}. Despite some game-changing achievements, several challenges in HAR remain unresolved.

\begin{itemize}
    \item \textbf{Noisy data} Sensory data inevitably includes lots of noise information on account of the inherent imperfections of sensors and often contains missing or even incorrect readings due to sensor failure~\cite{chen2021deep, guan2017ensembles}.
    \item \textbf{Intra-class variability} The same human activity, such as walking, may exhibit variations among different subjects or even for the same subject in different recording sessions~\cite{gholamiangonabadi2020deep}.
    \item \textbf{Inter-class similarity} Sensor data from different human activities may exhibit high similarities, which can make it challenging for a deep learning model to distinguish between them.~\cite{gholamiangonabadi2020deep}.
\end{itemize}

EnsembleLSTM~\cite{guan2017ensembles} has demonstrated the ability of ensemble methods to solve the problem of high variability in model output due to data quality. Multiple LSTM-based networks~\cite{hammerla2016deep} with the same structure, but different parameter values, are trained by using mini-batch and adjusting the size and position of the sliding window episode-wise. Then, a subset of the trained models is selected with the 'TopK' strategy. The final prediction is obtained by aggregating prediction of the selected models through the mode operation. 
While this method achieves good results, we hypothesize that it can be further improved by considering the following aspects. 

\textit{(i)}~A fundamental assumption of the bagging ensemble algorithm is that the trained models in the method are sufficiently independent of each other~\cite{breiman2001random}. Increasing the randomness of each model is a way to ensure this independence~\cite{breiman2001random}.  EnsembleLSTM~\cite{guan2017ensembles} introduces the random property by mini batch and varying parameters of sliding window, which can be further improved through sensor selection. Therefore, we hypothesize that a random selection of sensor \cite{breiman2001random} can further improve the final performance of the method. Also, due to the selection, the method would be less computational intensive by construction. 

\textit{(ii)}~The "TopK" strategy in EnsembleLSTM~\cite{guan2017ensembles} chooses the best performing models. However, this strategy does not consider if the selected models are sufficiently independent and lead to improvements when combined in an ensemble. Besides, the parameter $k$ in 'TopK' strategy is hard to initialize. By selecting model combinations with reinforcement learning rather than 'TopK', we hypothesize that we can further increase the ensemble performance and avoid the parameter initialization problem. 

Besides, it is worth noting that, having more sensor data does not necessarily lead to better performance~\cite{jensen2009more}. Instead, well-designed sensor selection can reduce the effect of intra-class variability and inter-class similarity. For instance, jogging and running may have similar vertical acceleration patterns but differ in horizontal acceleration. In such cases, removing certain sensor data can enhance the performance of the learning algorithm.



To address the challenges highlighted by the aforementioned observations, we propose a HAR framework called randomHAR. The framework combines sensor randomization and deep learning, while performing effective model selection through reinforcement learning.
Our contribution can be summarized as follows:
\begin{itemize}
    \item The proposed approach outperforms the state-of-the-art HAR ensemble method~\cite{guan2017ensembles} on six publicly available datasets.
    \item The proposed approach can avoid the parameter initialization hard problem in the state-of-the-art method.
    \item  The proposed approach can be applied to any HAR model that takes sensor signal as input.
    \item The PyTorch code of the proposed approach can be found in http://github for further study.
\end{itemize} 


\section{Related work}
Ensembles have been traditionally applied to decision trees as weak learners. The advantage over other classical models is that they react sensitive to e.g. different bootstraps or random feature selection~\cite{breiman2001random}. However, the work of~\cite{uddin2016random} still shows good performance of Random Forest on HAR. 
While machine learning-based approaches~\cite{nweke2018deep, lockhart2014limitations, ramasamy2018recent} have been successfully applied in the HAR domain over the past decade, they also face many challenges. For example, extracting complex features and recognizing time-series pattern.
The advances of deep learning have led to a point, where deep learning models achieve better results than other machine learning-based algorithms on many HAR benchmark datasets. 
Simple convolutional neural network (CNN) can be used to extract complex signal patterns. Alemayoh et al.~\cite{alemayoh2019deep} encodes the collected sensor signals into a 14x16 virtual image and then uses a CNN on the generated images to classify eight different activities. In \cite{chen2015deep}, CNNs are applied directly to the raw signal, while \cite{ronao2016human} combines the frequency domain convolution with time-domain convolution to achieve better performance. on the other hand, \cite{cheng2022real} tries to improve the performance by tuning hyperparameters on top of CNN, \cite{cruciani2020feature} by pretraining and \cite{nutter2018design} by employing PCA.
Compared with CNNs, LSTMs has received more attention in HAR with its ability to extract sequence information from raw signals. Chung et al.~\cite{chung2019sensor} starts to experiment with LSTM to solve simple classification problems, and Zhao et al.~\cite{zhao2018deep} proposes a bidirectional LSTM-based architecture to extract information from signals in both directions. The work of~\cite{guan2017ensembles} and~\cite{bock2021improving} have shown top performances and are thus used as references for our work. 
All methods mentioned above aim to reduce the high variance of the prediction results due to the strong noise of the HAR data and the variability of the subjects by optimizing the network parameters and architecture. In contrast, we try to solve the problem by optimizing an ensemble of LSTMs. 

\section{Methodology}
Given a HAR dataset $\mathcal{D} = \left\{s_0, \cdots,s_{n-1}\right\}$ where $n$ is the number of sensors containing in the dataset and $s_i, 0 \le i<n$ indicates the data collected with the specified sensor. The objective of a HAR task is to train a model $f$ on a specific subset $\mathcal{D}' \subset \mathcal{D}$ of the given data, such that, given any instance $x$ from the $\mathcal{D}/\mathcal{D}'$, $f(x) = y$, where $y$ is the ground-truth class of the instance $x$. 

\begin{figure}
    \centering
    \includegraphics[width=.47\textwidth]{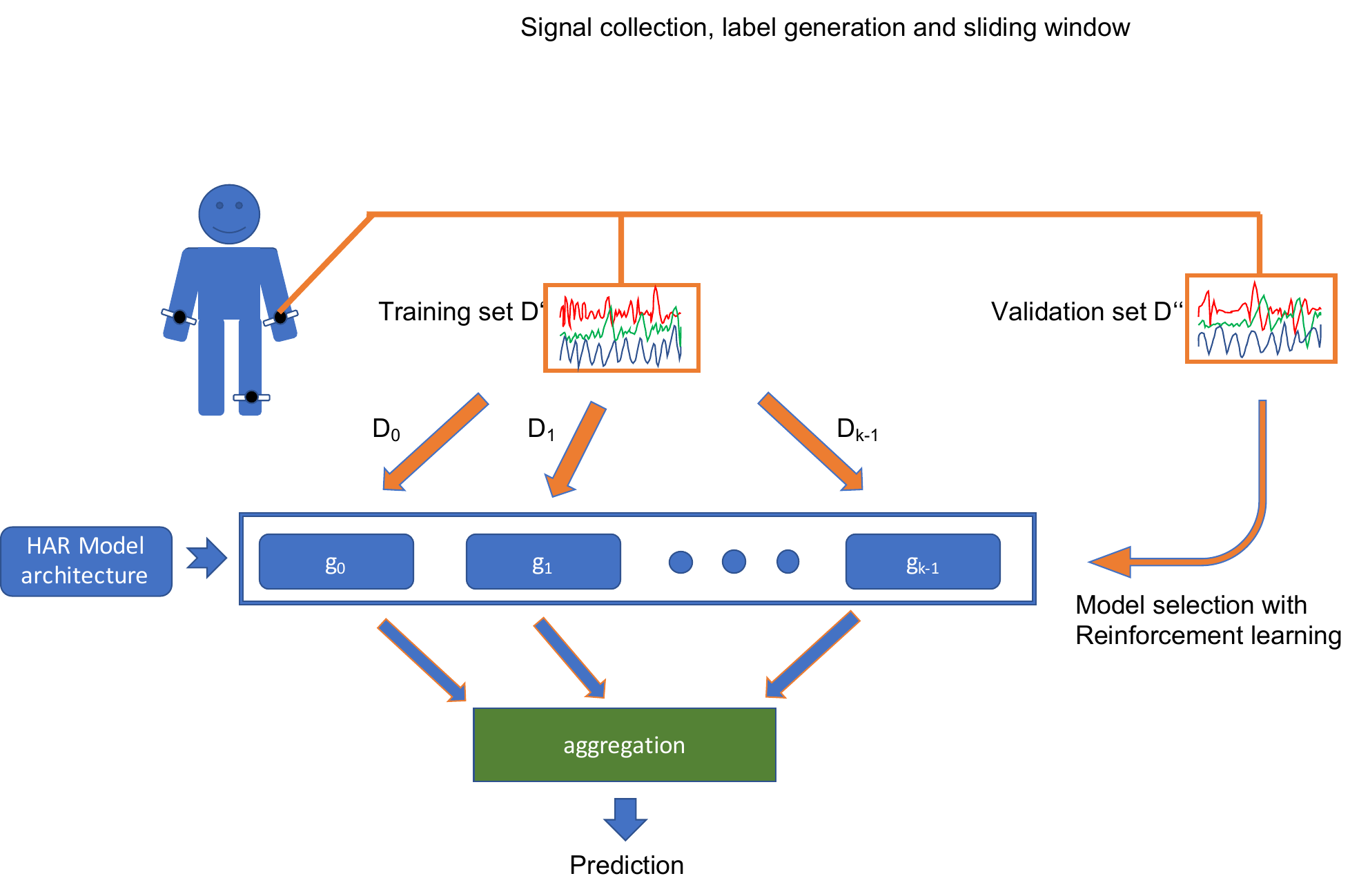}
    \caption{Framework of the proposed randomHAR algorithm.}
    \label{fig: framework}
\end{figure}

Figure~\ref{fig: framework} illustrates the pipeline of the proposed approach. It consists of two essential components, namely sensor selection and model selection. In the sensor selection stage, various subsets of the provided data are generated based on the included sensors. Several models with the same base model architecture are then trained on these generated subsets. In the model selection stage, a subset of the trained models is chosen. The output of these selected models are then aggregated using the 'mode' aggregation (i.e. majority voting) function to produce the final prediction.

\subsection{Sensor selection}
Given a training set $\mathcal{D}' = \left\{s_0, \cdots,s_{n-1}\right\}$ and a HAR model architecture. To generate subsets, $k$ binary vectors $m_i \in \left\{0, 1\right\}^{n}, 0 \le i < k$ are first generated by a Bernoulli distribution, where $k$ is the number of subsets to generate. These vectors are then utilized to generate $k$ subsets $\left\{\mathcal{D}_0, \cdots, \mathcal{D}_{k-1}\right\}$ by selectively incorporating only the sensor data whose corresponding value in the vector is 1. Each generated vector represents a unique subset. Each of these subsets is utilized to train a model with the given model architecture, resulting in the generation of $k$ models (classifiers) $\left\{g_0, \cdots, g_{k-1}\right\}$.

\begin{table*}[htbp]
    \centering
    \caption{Summary of the datasets used in the experiment. The abbreviations acc, gyro, mag denote 3d accelerometers, gyroscopes and magnetometers respectively.}
    \begin{tabular}{|c|c|c|c|c|c|}
        \hline
         Name & \#Subjects & Type & Sensors used & Freq(Hz) & Predicted classes\\
         \hline
         \multirow{3}{*}{DSADS} & \multirow{3}{*}{8} & \multirow{3}{*}{daily living} & \multirow{3}{*}{acc, gyro, mag}&\multirow{3}{*}{25Hz}&sitting, standing 1 and 2, walking 1, 2, 3, 4, 5 and 6,\\ &&&&& lying 1 and 2, running, exercising 1 and 2, \\ &&&&& cycling 1 and 2, rowing, jumping, playing basketball\\
         \hline
         \multirow{3}{*}{HAPT}&\multirow{3}{*}{30}&\multirow{3}{*}{daily living}&\multirow{3}{*}{acc, gyro}&\multirow{3}{*}{50Hz}&standing, sitting, lying, walking, walking upstairs, \\&&&&&walking downstairs, stand-to-sit, sit-to-stand, sit-to-lie,\\&&&&& lit-to-sit, stand-to-lie, lie-to-stand, null\\
         \hline
         \multirow{3}{*}{PAMAP2}&\multirow{3}{*}{9}&\multirow{3}{*}{daily living}&\multirow{3}{*}{acc, gyro}&\multirow{3}{*}{100Hz}&other, lying, sitting, standing, walking, running, cycling, \\&&&&& nordic walking, ascending stairs, descending stairs, \\&&&&&vacuum cleaning, ironing, rope jumping\\
         \hline
         RWHAR&15&daily living&acc&50Hz&jumping, lying, standing, sitting, running, walking, null\\
         \hline
         \multirow{4}{*}{SKODAR}&\multirow{4}{*}{1}&\multirow{4}{*}{car inspection}&\multirow{4}{*}{acc}&\multirow{4}{*}{98Hz}&write on notepad, open hood, close hood, \\&&&&&check front door gaps, open left front door,\\&&&&& close left front door, close both left door, check trunk gaps,\\&&&&& open and close trunk, check steering wheel, null\\
         \hline
         \multirow{3}{*}{OPPO}&\multirow{3}{*}{4}&\multirow{3}{*}{daily living}&\multirow{3}{*}{acc, gyro, mag}&\multirow{3}{*}{30Hz}&open/close door 1 and 2, fridge, dishwascher,\\&&&&& drawer 1,2 and 3 clean table, \\&&&&&drink from cup, toggle switch, null\\
         \hline
    \end{tabular}
    \label{tab: datasets}
\end{table*}

\subsection{Model selection}
\label{sec: model combination selection}
An episode-based reinforcement learning algorithm is applied to the generated models to find the best model combination. Finally, 'mode' aggregation (i.e. majority voting) is applied to obtain the final prediction.

The object function for the model combination task is modeled as
\begin{equation}
    J(\mu) = \int p_\mu(\theta)R(\theta) d\theta,\quad \mu^\star = \underset{\mu}{\operatorname{argmax}}\;J(\mu),
\end{equation}

where $p_\mu(\theta)= \mathcal{N}(\mu, I)$, is the search distribution to find the binary parameter vector $\theta = \left\{\theta^0, \cdots, \theta^k\right\}$ that maximizes reward 
\begin{equation}
\begin{split}
    R(\theta) &= \mathbb{E}_{x\sim \mathcal{D}''}\left[ \mathbbm{1}_{y, \text{ensemble}(\left\{\theta^0\cdot g_0(x), \cdots, \theta^k\cdot g_k(x) \right\})}\right]\\
    &\approx \frac{1}{N_1}\sum_{i=0}^{N_1}\left[\mathbbm{1}_{y, \text{ensemble}(\left\{\theta^0\cdot g_0(x_i), \cdots, \theta^k\cdot g_k(x_i) \right\})}\right] ,
\end{split}
\end{equation}
where $N_1$ is the number of samples generated to estimate the reward, $\mathcal{D}''$ is the validation set, $g$ represents the trained model (classifier) and $\theta$ represents the binary vector that used to select the classifiers for the final prediction. $\text{Ensemble}(\cdot)$ is the method to combine the predictions of the selected classifiers. We make use of the 'mode' aggregation function in the experiment.

The gradient of the objective function can be calculated through
\begin{equation}
\begin{split}
    \nabla J(\mu) &= \int R(\theta)\nabla_\mu \text{log}p_\mu(\theta)d\theta\\
    &\approx \frac{1}{N_2}\sum_{i=1}^{N_2}R(\theta_i)\nabla_\mu \text{log} p_w(\theta_i)\\
    &\approx \frac{1}{N_2}\sum_{i=1}^{N_2}R(\theta_i)(\theta_i - \mu),
\end{split}
\end{equation}
where $N_2$ is the number of samples in Monte Carlo estimation.


\section{Experiment}
In this section, we design experiments to investigate various essential questions pertaining to the proposed approach: \textit{(i)}~Whether the proposed approach can outperform the state-of-the-art ensemble method, ensembleLSTM~\cite{guan2017ensembles}? \textit{(ii)}~Whether the reinforcement learning-based model selection process in the proposed approach necessary? 
\textit{(iii)}~how generalized the proposed approach is?


\subsection{Experiment setting}
\begin{table*}[htbp]
    \centering
    \caption{Performance of the proposed approach compared with the ensembleLSTM method on six datasets.}
    \begin{tabular}{|c|c|c|c|c|}
        \hline
        Dataset& base & ensembleLSTM & randomHAR-all& randomHAR-rl  \\
        \hline
        DSADS&0.856 $\pm$ 0.007&0.865 $\pm$ 0.006 & 0.866 $\pm$ 0.005&\textbf{0.874 $\pm$ 0.001}\\
        \hline
        HAPT&0.795 $\pm$ 0.015 &0.802 $\pm$ 0.015 & 0.805 $\pm$ 0.007 &\textbf{0.830 $\pm$ 0.010}\\
        \hline
        PAMAP2&0.735 $\pm$ 0.012 &0.760 $\pm$ 0.011 & 0.756 $\pm$ 0.008 &\textbf{0.762 $\pm$ 0.008} \\
        \hline
        RWHAR& 0.718 $\pm$ 0.021&0.733 $\pm$ 0.004 & 0.740 $\pm$ 0.004 & \textbf{0.751 $\pm$ 0.010}\\
        \hline
        SKODAR&0.972 $\pm$ 0.017 &0.992 $\pm$ 0.006 & 0.992 $\pm$ 0.004 & \textbf{0.994 $\pm$ 0.002} \\
        \hline
        OPPO&0.373 $\pm$ 0.007 &0.371 $\pm$ 0.002 & 0.370 $\pm$ 0.002 & \textbf{0.380 $\pm$ 0.005}\\
        \hline
    \end{tabular}
    \label{tab: performance}
\end{table*}
In the first two experiments, we use the deepConvLSTM variant model proposed in the ISWC 2021 best paper~\cite{bock2021improving} as the base model and followed most of the experimental settings in that paper. In the last experiment, we utilize the deep convolutional neural network (CNN) as the base model to test the generality of the proposed approach. 

We evaluated the model on six publicly available datasets, namely a preprocessed version of the Opportunity (OPPO) dataset~\cite{roggen2010collecting} as well as five popular HAR datasets namely DSADS~\cite{barshan2014recognizing}, HAPT~\cite{reyes2015smartphone}, PAMAP2~\cite{reiss2012introducing}, RealWorld HAR~(RWHAR)~\cite{sztyler2016body} and SKODAR~\cite{zappi2012network}. The descriptions of these datasets are summarized in Table~\ref{tab: datasets}. 
In addition, F1-score (macro) and Leave-One-Subject-Out (LOSO) Cross-Validation method are applied on all the datasets (except SKODAR) to assess the performance of the model. The SKODAR dataset contains only one subject, and the 5-folds Cross-Validation is applied to this dataset. 

All experiments are repeated 5 times to make the result more plausible. We use Adam optimizer with the initial learning rate 1e-4 to train the model. The learning rate decays by a factor of 10 with patience equal to 5 epochs. The maximal training epochs is 50 with early stopping, which patience is set to 10. For the model combination selection, the parameter $N_1$ is set to the size of the validation set and $N_2$ is set to 10. 

When training the ensembleHAR model in our experiments, instead of using the episode-wise bagging, we take the data utilized in each epoch (after injecting the randomness) to train an individual model. By adopting this approach, each variant of the model is fully trained at the expense of increased resource usage. Simultaneously, it improves the comparability of the method with the proposed approach, while preserving its fundamental principle. The performances of the trained method are compared to that in the corresponding paper~\cite{guan2017ensembles} and no degradation in performance was observed.

\subsection{Evaluation and discussion}

To assess the efficacy of the proposed approach, a comparative analysis of the following four distinct methods are conducted: \textit{(i)}~The base deepConvLSTM model (base), \textit{(ii)}~The deepConvLSTM model with the ensemble method proposed in \cite{guan2017ensembles} (ensembleLSTM), \textit{(iii)}~The deepConvLSTM with the proposed ensemble approach without trained model selection (randomHAR-all). \textit{(iv)}~The deepConvLSTM model with the proposed ensemble approach with reinforcement learning model selection (randomHAR-rl).
Ten models are trained in both ensemble methods. For the 'ensembleLSTM' method, five of generated models are utilized for the final prediction.

We summarize the result in Table~\ref{tab: performance}. We can see that 'randomHAR-rl' achieves the best performance on all datasets. Among them, by employing significance test (with 'scipy' package), 'RandomHAR-rl' is shown to significantly outperform other methods on four out of six datasets, namely DSADS, HAPT, RWHAR, OPPO. It reduces the variance while achieving improvement in F1 scores. This confirms our hypothesis: strategies to increase the randomness of the trained model (sensor selection) and to improve the quality of the model selection (reinforcement learning) can lead to enhanced performance. When comparing 'ensembleLSTM' and 'randomHAR-all', we found that 'randomHAR-all' does not exhibit any significant advantage in terms of results. This indicates that model selection is imperative in the proposed approach. This observation is reasonable, since an inappropriate sensor selection can severely limit the performance of the corresponding trained model. 

\begin{figure}
    \centering
    \includegraphics[width = .4\textwidth]{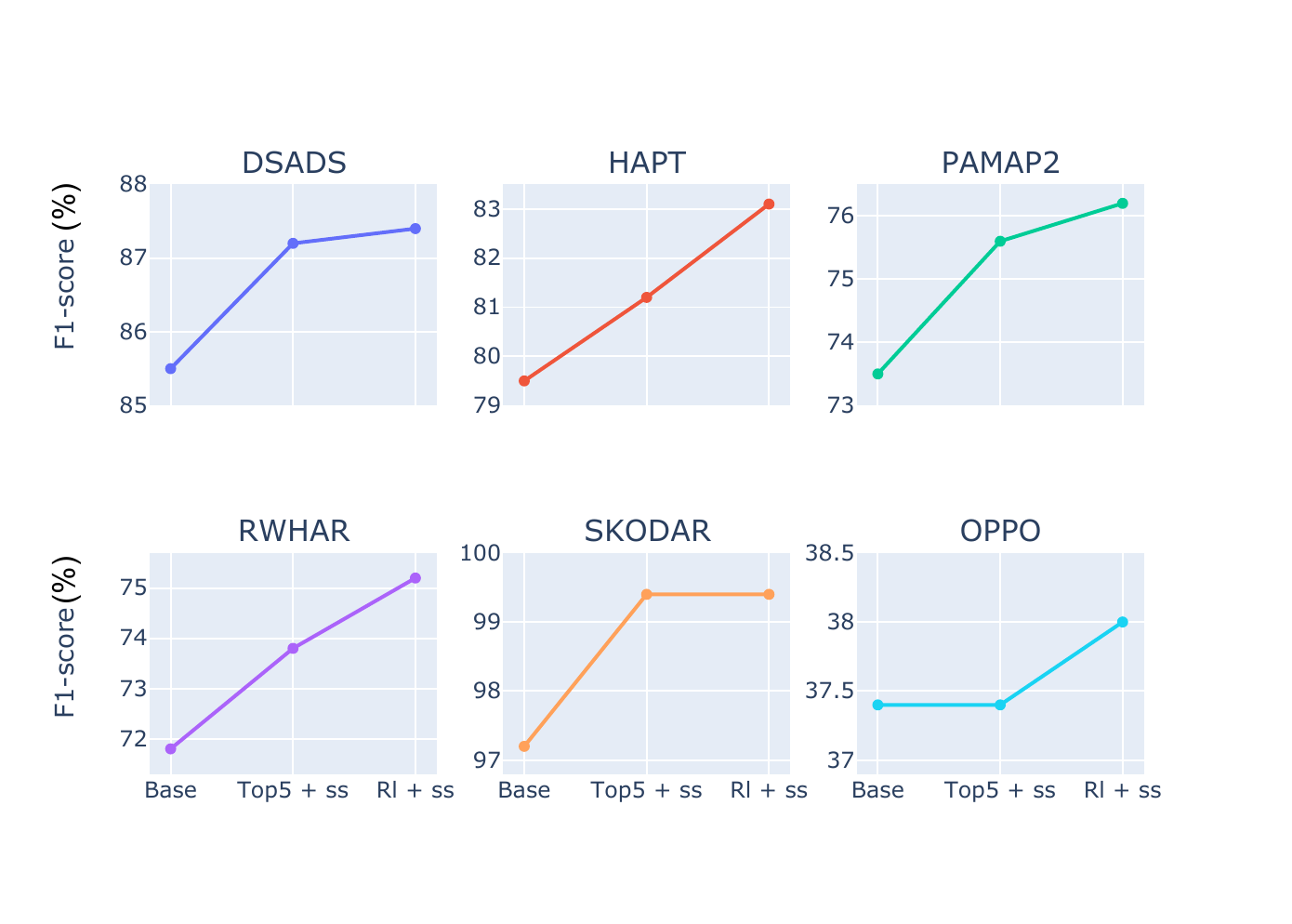}
    \caption{RandomHAR Performance with and without reinforcement learning model selection strategy.}
    \label{fig: ablation}
\end{figure}

To evaluate the effectiveness of the reinforcement learning model selection process, we compared the performance of RandomHAR using the "TopK" strategy (Top5 + ss) with that of the RandomHAR method using the reinforcement learning strategy (Rl + ss). According to the result in Fig.~\ref{fig: ablation}, the reinforcement learning strategy brings performance gains on all datasets. while it should be noted that the improvement brought by the reinforcement learning strategy on three of the datasets is not significant enough. Nevertheless, since reinforcement learning can learn the optimal number of models to be selected, the reinforcement learning strategy avoids the need for hyperparameter initialization, which is a challenging problem in the "TopK" strategy.

To demonstrate the generality of the proposed approach, we conducted an experiment wherein the base model in randomHAR was replaced with a CNN model\footnote{The architecture of the model can be found in http://github}. The result is summarized in TABLE~\ref{tab: cnn model}. As shown, the proposed approach leads to improved performance across all datasets, although the overall performance is not as good as that obtained using the ConvLSTM based model. This finding provides evidence for the generality of the proposed approach, and indicate its potential applicability to other base models.
\begin{table}[]
    \centering
    \caption{Performance of the proposed approach using CNN model as base model.}
    \begin{tabular}{|c|c|c|}
        \hline
        Dataset& base & randomHAR  \\
        \hline
        DSADS & 0.754 & \textbf{0.776} \\
        \hline
        HAPT & 0.713 & \textbf{0.770}\\
        \hline
        PAMAP2 & 0.588 & \textbf{0.626} \\
        \hline
        RWHAR & 0.686 & \textbf{0.72}\\
        \hline
        OPPO & 0.305 & \textbf{0.367}\\
        \hline
    \end{tabular}
    \label{tab: cnn model}
\end{table}

Besides, Random forest and neural network are currently widely employed in various fields. An intuitive question is why not just utilize them to generate the final decision using the predictions of all the trained models as input. To address this concern, we explore the efficacy of Multilayer Perceptron (MLP), as a substitute for model selection, to generate the final prediction. However, the result reveals that the performance of the MLP is unstable and inferior compared to the proposed approach, which may be due to over-fitting.

\section{Conclusion and future work} 
In this paper, we propose a novel ensemble approach for HAR and improve the performance of existed HAR Ensemble methods. This is achieved by increasing the randomness of individual models in the trained model set and improving the model combination selection strategy. 
Although the proposed approach yields promising results, there are still many avenues for further research. For example, \textit{(i)}~can we optimize the reward function for the model selection process, e.g., can we accelerate the convergence by subtracting the average reward of the old agent? \textit{(ii)}~instead of randomly selecting sensors, can meta-features be utilized to enhance the targeted performance of the trained models? \textit{(iii)}~can the original HAR problem be split into two consecutive sub-problems, and multiple models trained on biased datasets be combined to make predictions?  %

\bibliography{2023_wimob}
\bibliographystyle{IEEEtran}

\end{document}